\documentclass[10pt,twocolumn,letterpaper]{article}

\usepackage{cvpr}              %

\usepackage[dvipsnames]{xcolor}

\usepackage{float}

\definecolor{cvprblue}{rgb}{0.21,0.49,0.74}
\usepackage[pagebackref,breaklinks,colorlinks,citecolor=cvprblue]{hyperref}

\def\paperID{} %
\def\confName{CVPR}
\def\confYear{2024}

\title{3D Representation Methods: A Survey}

\author{Zhengren Wang \\
Peking University\\
{\tt\small wzr@stu.pku.edu.cn}
}

\begin{document}
\maketitle
\begin{abstract}
The field of 3D representation has experienced significant advancements, driven by the increasing demand for high-fidelity 3D models in various applications such as computer graphics, virtual reality, and autonomous systems.
This review examines the development and current state of 3D representation methods, highlighting their research trajectories, innovations, strength and weakness.
Key techniques such as Voxel Grid, Point Cloud, Mesh, Signed Distance Function (SDF), Neural Radiance Field (NeRF), 3D Gaussian Splatting, Tri-Plane, and Deep Marching Tetrahedra (DMTet) are reviewed.
The review also introduces essential datasets that have been pivotal in advancing the field, highlighting their characteristics and impact on research progress. Finally, we explore potential research directions that hold promise for further expanding the capabilities and applications of 3D representation methods.
\end{abstract}

\section{Introduction}
\label{sec:intro}
The field of 3D representation has experienced substantial growth, spurred by the advances in deep learning and its critical applications across diverse domains. 3D representation forms the backbone of many technological advancements. In computer graphics, it allows for the creation of realistic models and environments, enhancing the visual fidelity and immersive experience in both entertainment and simulation. Virtual reality relies heavily on accurate 3D representations to create lifelike experiences that can be used for gaming, training, and education. In autonomous navigation, 3D models of the environment are essential for accurate mapping and obstacle detection, ensuring safe and efficient movement of autonomous vehicles. Medical imaging leverages 3D representations to provide detailed views of anatomical structures, aiding in diagnosis and surgical planning.

Given the wide-ranging impact of 3D representation technologies, it is imperative to understand the underlying methods and advancements in this area. This review aims to explore the development and current state of 3D representation methods, summarize key techniques, introduce significant datasets, and identify promising directions for future research.

\subsection{Research Focus and Development Process}
The evolution of 3D representation methods has been marked by a progression from basic geometric constructs to sophisticated, data-driven models. Early approaches focused on geometric primitives and spatial data structures. Over time, with advancements in computing power and machine learning, more complex and flexible representations emerged.
\begin{itemize}
    \item Early Methods and Geometric Models: Initial efforts in 3D representation relied heavily on explicit geometric descriptions, such as polygonal meshes and constructive solid geometry (CSG). These methods were foundational in computer-aided design (CAD) and early computer graphics.

    \item Volume-based Representations: The introduction of voxel grids offered a discrete representation of 3D space, suitable for medical imaging and volumetric data processing. However, voxel grids suffer from high memory usage and resolution limitations.

    \item Point Clouds: Point clouds, which represent objects as a collection of discrete points in space, became popular with the advent of LiDAR and depth sensors. While efficient for capturing raw spatial data, point clouds lack connectivity information, posing challenges for downstream processing.

    \item Mesh-based Methods: Polygonal meshes, composed of vertices, edges, and faces, provide a compact and versatile representation. They are widely used in computer graphics and animation. Mesh-based methods have evolved to support dynamic and deformable models.

    \item Implicit Representations: Techniques such as Signed Distance Functions (SDFs) offer a continuous and implicit way to describe surfaces. SDFs are particularly useful for collision detection and shape analysis.

    \item Neural Representations: Recent advancements have seen the rise of neural network-based representations, such as Neural Radiance Fields (NeRFs), which encode 3D scenes using neural networks. These methods have revolutionized 3D reconstruction and novel view synthesis.

    \item Hybrid Methods: Combining multiple representation methods, such as DMTet, Tri-plane, and 3D Gaussian Splatting, can leverage the strengths of each approach to handle complex scenarios more effectively.
\end{itemize}

\section{3D Representation Methods}
In this section, we present an overview of the paramount 3D representation techniques, encompassing, but not restricted to, the following approaches:
\subsection{Voxel Grid}
Voxel grid representation is a method for modeling 3D objects where the space is divided into a regular grid of cubes, known as voxels (volume elements). Each voxel in the grid can store information such as color, density, or material properties, allowing for a detailed volumetric representation of the object. This technique is particularly advantageous for representing complex geometries and internal structures that might be challenging to capture with surface-based methods like meshes.
One of the foundational advantages of voxel grids is their straightforward implementation and intuitive understanding, as they extend the concept of 2D pixels to three dimensions. This makes voxel grids particularly useful in applications where volumetric data is naturally produced, such as medical imaging (e.g., MRI and CT scans), where organs and tissues are represented as a 3D array of voxels.
Voxel grids are pivotal in computer graphics and game development, exemplified by their use in destructible environments and procedural generation in games like 'Minecraft'. 
These applications leverage the simplicity and flexibility of voxel grids to create and manipulate 3D environments dynamically.

One notable work is by Maturana and Scherer  \cite{maturana2015voxnet} with their development of VoxNet, a 3D convolutional neural network (CNN) that operates directly on voxel grids for object recognition tasks. This approach demonstrated the potential of combining voxel representations with deep learning techniques to achieve state-of-the-art performance in 3D shape classification and recognition tasks.
Building on the foundation of utilizing voxel grids in deep learning, the work on VoxGRAF by Schwarz et al. \cite{schwarz2022voxgraf} introduces a novel approach to 3D-aware image synthesis using sparse voxel grids. Their research demonstrates the feasibility of substituting monolithic multilayer perceptrons (MLPs) with 3D convolutions by effectively combining sparse voxel grids with sophisticated techniques like progressive growing, free space pruning, and appropriate regularization. This methodology not only ensures efficient rendering from arbitrary viewpoints but also guarantees 3D consistency and high visual fidelity in the synthesized images, marking a significant leap in the domain of image synthesis and rendering. Further enriching the applications of voxel grids, Vox-E by Sella et al. \cite{sella2023vox} ventures into the realm of voxel editing of 3D objects guided by textual descriptions. This innovative work introduces a volumetric regularization loss that directly operates in 3D space, leveraging the explicit nature of voxel representation to enforce a correlation between the global structures of the original and edited objects. This advance opens new vistas in the domain of 3D modeling and editing, where detailed and intensive modifications can be intuitively guided through natural language inputs, enhancing both the accessibility and flexibility of 3D object manipulations. 
Complementing the foregoing advancements, 
Tatarchenko et al. \cite{tatarchenko2017octree} presented a deep convolutional decoder architecture that can generate volumetric 3D outputs in a compute- and memory-efficient manner by using an octree representation, which allows representing much higher resolution
outputs with a limited memory budget.

\begin{figure}[H]
    \centering
    \includegraphics[width=0.9\linewidth]{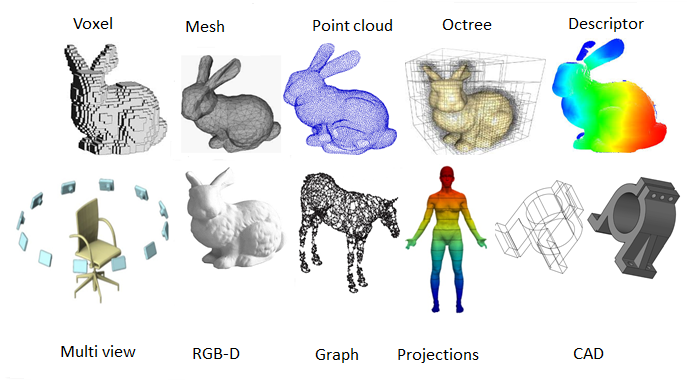}
    \caption{Examples of different 3D data representations. Figure from \cite{gezawa2020review} }
    \label{fig:3d}
\end{figure}
\subsection{Point Cloud}
Point cloud representation is a fundamental method for modeling and understanding three-dimensional (3D) structures. A point cloud is essentially a set of data points in space, which represents the external surface of an object or a scene. These points are typically defined by their coordinates in a 3D space (x, y, z) and can sometimes include additional information such as color, intensity, and normal vectors. Point clouds are generated by 3D scanning technologies like LiDAR (Light Detection and Ranging), photogrammetry, and depth cameras, which capture the shape of objects and environments with high precision. One of the key advantages of point clouds is their simplicity and direct representation of the 3D world. Unlike other 3D representations such as meshes or voxel grids, point clouds do not require connectivity information or volumetric data, making them easier to generate and manipulate. However, this simplicity also comes with challenges. Point clouds are often sparse, unstructured, and can contain noise, making processing tasks like segmentation, recognition, and reconstruction more complex.

A landmark work in this domain is PointNet by Qi et al. \cite{qi2017pointnet}. PointNet proposed a novel neural network architecture that directly consumes point clouds, overcoming the need for converting them into structured forms like voxel grids or meshes. This work demonstrated that deep learning models could achieve impressive performance on 3D recognition tasks by respecting the permutation invariance of point sets and using a series of shared multi-layer perceptrons (MLPs) and max pooling to aggregate global features.
PointNet++ \cite{qi2017pointnet++} extended the original PointNet by incorporating hierarchical learning, allowing the network to capture local structures at multiple scales. This approach significantly improved the ability to handle varying point densities and capture fine-grained geometric details.
Another influential work is Dynamic Graph CNN (DGCNN) by Wang et al. \cite{wang2019dynamic}. This method builds graphs dynamically in each layer of the network, connecting neighboring points to capture local geometric relationships more effectively. DGCNN has been shown to perform exceptionally well on various tasks, including classification, segmentation, and part segmentation, by leveraging the flexibility of graph-based representations.
In addition to these methods, recent research has explored transformer-based architectures for point clouds. Point Transformer by Zhao et al. \cite{zhao2021point} is a notable example, adapting the transformer architecture to capture both local and global dependencies in point clouds. This approach demonstrated state-of-the-art performance in tasks like classification and segmentation by effectively modeling the complex interactions between points.
Notably, Point-E by Nichol et al. \cite{nichol2022point}, the state-of-the-art 3D diffusion-based point cloud generative model, first generates a single synthetic view using a text-to-image
diffusion model, and then produces a 3D point cloud using a second diffusion model which conditions on the generated image for generating 3D point clouds from complex prompts.

\subsection{Mesh}
Mesh representation is a fundamental approach in 3D computer graphics and computational geometry for representing the surface of a 3D object. It is composed of vertices, edges, and faces that form a polyhedral shape. The vertices are point clouds in 3D space, the edges connect pairs of vertices, and the faces (typically triangles or quadrilaterals) are defined by edges connecting three or more vertices. Meshes are popular because they offer a good balance between simplicity and expressive power. They can approximate complex geometries with arbitrary precision by adjusting the number of vertices and faces. Additionally, meshes are well-supported by graphics hardware, making them efficient for rendering and simulation tasks. The flexibility of mesh representation allows for efficient computation of surface properties, such as normals and curvature, which are crucial for realistic rendering and physical simulations.

Neural 3D Mesh Renderer by Kato et al. \cite{kato2018neural}, which introduces a differentiable renderer that allows for end-to-end training of neural networks involving mesh representations. This differentiability enables the direct optimization of 3D mesh parameters through image-based loss functions, bridging the gap between 2D image processing and 3D geometry manipulation in neural networks.
Pixel2Mesh by Wang et al. \cite{wang2018pixel2mesh} proposes an end-to-end network for reconstructing 3D meshes from single RGB images. The approach integrates a graph-based convolutional network with a mesh deformation framework, enabling the progressive refinement of an initial ellipsoid into a detailed 3D mesh. This method demonstrates the capability of deep learning to generate accurate and detailed 3D shapes from 2D inputs.
AtlasNet by Groueix et al. \cite{groueix2018papier} introduces a framework for generating 3D meshes using deep neural networks. AtlasNet represents a 3D surface as a collection of parametric surface patches, which are learned and assembled to form the final mesh. This approach offers flexibility in capturing complex geometries and facilitates efficient learning and inference.
Mesh R-CNN by Gkioxari et al. \cite{gkioxari2019mesh} extends the Mask R-CNN framework to handle 3D mesh prediction. It combines image-based feature extraction with graph convolutional networks to predict the 3D shape of objects within images, achieving state-of-the-art results in 3D shape reconstruction tasks. The integration of image segmentation and 3D reconstruction in a single framework exemplifies the advancements in leveraging mesh representations for complex scene understanding.

\subsection{Signed Distance Function (SDF)}
The Signed Distance Function (SDF) representation is widely used for representing the water-tight shapes and surfaces of 3D objects. An SDF encodes the geometry of an object by defining the distance from any point in space to the nearest surface of the object, with the sign indicating whether the point is inside (negative sign) or outside (positive sign) the object. This representation allows for a smooth and continuous description of surfaces. One of the foundational works in the field is the use of SDFs in implicit surface modeling, where surfaces are defined by the zero level set of the SDF. This approach allows for the seamless handling of topological changes, such as merging and splitting of surfaces, which is particularly advantageous in simulations involving fluid dynamics and deformable objects.

A seminal work that propelled the use of SDFs in 3D reconstruction is DeepSDF by Park et al. \cite{park2019deepsdf}. This work introduced to learn continuous SDFs directly from raw data, enabling high-fidelity shape representation and reconstruction from incomplete and noisy observations. The authors demonstrated that DeepSDF could capture fine details and complex topologies, outperforming traditional mesh-based and voxel-based methods in terms of accuracy and efficiency.
For articulated shape representation, Mu et al. \cite{mu2021sdf} introduce Articulated SDFs (A-SDFs) to model articulated shapes with disentangled latent spaces for shape and articulation. Their method enhances the representational power of SDFs by incorporating the ability to adjust models at test time through Test-Time Adaptation. This work underscores the versatility and adaptability of SDF-based methods in capturing the nuances of articulated and deformable objects, extending the utility of SDFs beyond static shapes to dynamic models. 
Mittal et al. \cite{mittal2022autosdf} introduced AutoSDF, which relies on autoregressive shape priors for effective 3D shape completion, reconstruction, and generation. This work stands out by demonstrating an ability to handle multimodal 3D tasks while outperforming state-of-the-art methods optimized for singular tasks. This approach not only solidifies the utility of SDFs in capturing complex geometries but also showcases the power of deep learning in extracting and utilizing shape priors from vast datasets. Contrast to AutoSDF, the SDF-StyleGAN by Zheng et al. \cite{zheng2022sdf} extends the capabilities of StyleGAN2 for 3D shape generation using implicit SDF as the shape representation. This novel integration addresses the challenge of high-quality 3D shape geometry generation by incorporating specialized discriminators for real and fake SDF values and gradients, hence significantly lifting the visual quality and geometric accuracy of generated shapes. LAS-Diffusion \cite{zheng2023locally} highlights a diffusion-based framework tempered with a novel view-aware local attention mechanism, emphasizing the improvement in local controllability and generalizability of 3D shapes. The method effectively utilizes 2D sketch images as inputs, opening new avenues for image-conditioned 3D shape generation that can cater to detailed specifications and artistic vision. 
Notably, Shap-E by Jun et al. \cite{jun2023shap} from OpenAI, directly generates the parameters of implicit functions (SDF and NeRF). This approach offers a faster convergence and achieves comparable or superior sample quality despite tackling a higher-dimensional, multi-representation output space.

\subsection{Neural Radiance Field (NeRF)}
Neural Radiance Field (NeRF) has emerged as a transformative approach in the domain of 3D representation and view synthesis. This method encodes the volumetric scene representation in a neural network that can synthesize novel views of complex scenes from a sparse set of input images. NeRF, by Mildenhall et al. \cite{mildenhall2021nerf}, utilizes a fully connected deep neural network to predict the color and density of a point in space based on its spatial coordinates and viewing direction. This allows the model to render photorealistic images from novel viewpoints by integrating these predictions along rays cast through the scene. The impressive quality of NeRF's rendered images has established it as a foundational technique in 3D computer vision and graphics.

The NeRF methodology has been extended and refined in several important works. For instance, GRAF by Schwarz et al. \cite{schwarz2020graf} explores how generative models can be combined with radiance fields to synthesize 3D-consistent images. GRAF leverages the power of generative adversarial networks (GANs) to produce high-quality images that are coherent from different viewpoints, thereby advancing the intersection of 3D representation and image synthesis.
NeRF++ by Zhang et al. \cite{zhang2020nerf++} investigates the limitations of NeRF, particularly in handling unbounded scenes, and proposes extensions that improve its representation of such environments. NeRF++ enhances the model’s capability to render scenes with complex geometry and varying depth, broadening the applicability of NeRF techniques.
For scenarios where camera parameters are unknown, NeRF-- by Wang et al. \cite{wang2021nerf} presents a method to optimize both the scene representation and the camera parameters simultaneously. This innovation allows NeRF to be used in more flexible and less controlled environments, enhancing its practicality for real-world applications.
NeRF-W by Martin-Brualla et al. \cite{martin2021nerf} also extends the applicability of NeRF to unconstrained and diverse photo collections. This work addresses the challenges posed by real-world image datasets, which often include variations in lighting, weather, and occlusions. By incorporating strategies to handle these variations, NeRF in the Wild demonstrates the robustness of NeRF models in synthesizing high-quality views from less controlled image data.
Mip-NeRF by Barron et al. \cite{barron2021mip} addresses the challenge of aliasing artifacts in NeRF via introducing a multiscale representation. This approach enables more robust and accurate renderings, especially at varying levels of detail and scale, thereby improving the visual quality of the synthesized images.
InstantNGP by Müller et al. \cite{muller2022instant} focuses on optimizing the efficiency of NeRF training and inference. By leveraging a combination of hash table-based encodings and efficient GPU computations, Instant NGP dramatically reduces the time required to train NeRF models, making the technology more accessible for real-time applications.
Mip-NeRF 360 by Barron et al. \cite{barron2022mip}, introduces a multiscale representation that effectively handles anti-aliasing and the complexities of infinite scenes, providing more accurate and visually pleasing renderings.
Plenoxels by Fridovich et al. \cite{fridovich2022plenoxels} represents a significant departure from the neural network-based methods typical of NeRF. Plenoxels utilize a sparse voxel grid to directly encode the radiance field, bypassing the need for neural networks entirely. This method leverages the simplicity and efficiency of voxel grids while still achieving high-quality view synthesis.
Similarly, Zip-NeRF by Barron \cite{barron2023zip} focuses on reducing the computational overhead associated with NeRF by employing a grid-based structure that allows for efficient sampling and representation of the radiance fields. This approach not only speeds up the rendering process but also mitigates aliasing artifacts, resulting in sharper and more precise image outputs.

\subsection{3D Gaussian Splatting}
3D Gaussian Splatting (3DGS) is an innovative representation technique for modeling and rendering 3D scenes using Gaussian distributions. The core idea behind 3DGS is to represent the scene as a collection of 3D Gaussians, which can be used to approximate the geometry and appearance of the scene efficiently. Each Gaussian in this representation encapsulates spatial and color information, allowing for smooth and continuous approximations of surfaces and textures.
The representation offers several advantages over traditional NeRF-based models. Firstly, it allows for real-time rendering, which is crucial for applications in virtual reality (VR) and augmented reality (AR). The real-time capability is achieved by leveraging the continuous nature of Gaussians, which can be rendered efficiently using point-based techniques and optimized through advanced acceleration structures.

The seminal work by Kerbl et al. \cite{kerbl20233d} introduces the concept of using 3D Gaussian splats to represent radiance fields, enabling real-time rendering of complex scenes. The method demonstrates how Gaussian splats can be efficiently rendered and integrated into real-time applications, providing a substantial improvement in performance compared to traditional NeRF and volumetric rendering.
DreamGaussian \cite{tang2023dreamgaussian} by Zhang et al. and Gaussiandreamer by Yi et al. \cite{yi2023gaussiandreamer} extends the usage of 3DGS by incorporating generative models to create 3D content. DreamGaussian and Gaussiandreamer leverage the efficiency of Gaussian splats to enable the creation of high-quality 3D content with reduced computational overhead.
The generative approach allows for the synthesis of new scenes and objects, providing a powerful tool for content creators in games and movies. 
Mip-Splatting by Fanello et al. \cite{yu2023mip} addresses the issue of aliasing in 3DGS. Aliasing occurs when the resolution of the representation is insufficient to capture the details of the scene, leading to visual artifacts. Mip-Splatting introduces a hierarchical representation that mitigates aliasing by using multiple levels of detail. This approach ensures that the rendering remains smooth and detailed across different scales.
Dynamic scenes pose additional challenges \cite{wu20234d}, Du et al. \cite{yang2023deformable} presents a method for reconstructing dynamic scenes using deformable 3D Gaussians. The approach allows for the accurate modeling of moving objects and changes in the scene, providing high-fidelity reconstructions from monocular video input. 
Similarly, Gao et al. \cite{luiten2023dynamic} explores the use of dynamic 3D Gaussians for tracking objects in a scene. The technique focuses on persistent dynamic view synthesis, which enables continuous and coherent tracking of objects over time. 

\subsection{Hybrid Methods}
\subsubsection{Deep Marching Tetrahedra (DMTet)}
Deep Marching Tetrahedra (DMTet) by Shen et al. \cite{shen2021deep} is a sophisticated technique for 3D representation that builds upon the concept of marching tetrahedra, a method used for isosurface extraction. Unlike traditional marching cubes or marching tetrahedra, which operate on a fixed grid, DMTet leverages deep learning to enhance the flexibility and accuracy of the representation.
At the core of DMTet is the use of a neural network to predict the occupancy and appearance of the 3D space. This network typically takes a latent code or an implicit function as input, which encodes the shape and structure of the object. The space is discretized into tetrahedra, small, tetrahedron-shaped volumetric elements. Each tetrahedron's vertices are evaluated to determine their occupancy values, indicating whether they are inside or outside the surface of the object. These occupancy values are then used to extract the surface by interpolating the values at the tetrahedron's vertices, similar to how traditional marching methods operate but with more adaptive and learned boundaries.

One significant advantage of DMTet is its ability to handle complex topologies and fine details in 3D shapes. The learned representation allows for adaptive resolution, where the network can allocate more tetrahedra to regions requiring higher detail and fewer tetrahedra to simpler regions. This adaptive meshing ensures efficient use of computational resources while maintaining high fidelity in the represented shapes.
Furthermore, DMTet integrates seamlessly with differentiable rendering techniques, allowing for end-to-end optimization of the 3D shape based on image-based loss functions. This integration means that the network can be trained directly from 2D images, improving the 3D representation through backpropagation of errors from rendered images, leading to highly detailed and accurate 3D models.
While DMTet focuses on the geometry aspect of 3D representation, it is also crucial to acknowledge the role of appearance in creating realistic and high-quality 3D content. For instance, Fantasia3D by Chen et al. \cite{chen2023fantasia3d} is introduced as a novel method capable of disentangling geometry and appearance, where a mesh is optimized from scratch with DMTet. Fantasia3D's key contribution lies in its ability to facilitate the separate manipulation of geometric and appearance attributes, enabling more detailed and control over the final 3D model. For another example, Magic3D by Lin et al. \cite{lin2023magic3d} has improved qualitative results by adopting two-stage training where a course NeRF is first learned and then converted to a DMTet representation for further optimization.
\subsubsection{Tri-plane}
Tri-plane \cite{chan2022efficient} approximately divides the 3D space into three orthogonal planes (typically aligned with the X, Y, and Z axes), allowing for efficient encoding and rendering of 3D information. Each plane captures different aspects of the spatial structure, and together, they provide a comprehensive representation. The tri-plane representation leverages the fact that complex 3D structures can often be decomposed into simpler 2D projections, which can then be recombined to form a complete 3D model. This decomposition simplifies the processing and manipulation of 3D data by reducing the dimensionality of the problem from three dimensions to two. In practice, this means that the information about the 3D object or scene is encoded in three separate 2D grids or textures, each corresponding to one of the orthogonal planes. These grids can store various attributes, such as color, depth, or occupancy, which can then be processed using standard 2D image processing techniques.
One of the key advantages of tri-plane representation is its ability to balance computational efficiency with the quality of the 3D reconstruction. By leveraging the inherent structure of 3D objects and scenes, this method can achieve high levels of detail and accuracy without the need for complex and resource-intensive 3D volumetric processing. This makes it particularly suitable for real-time applications, such as virtual reality (VR) and augmented reality (AR), where performance is critical.

Recent studies have leveraged the utility of tri-plane representations across various applications. Huang et al. \cite{huang2023tri} introduced a tri-perspective view (TPV) representation, adding two additional planes perpendicular to the bird's eye view (BEV), demonstrating that camera inputs alone can match the performance of LiDAR-based methods in the LiDAR segmentation task on nuScenes. This achievement underscores the versatility and efficiency of tri-plane representations in enhancing vision-based 3D semantic occupancy prediction. In another significant contribution, OTAvatar by Ma et al.   \cite{ma2023otavatar}, which utilizes a generalized controllable tri-plane rendering solution. Remarkably, this method enables the construction of personalized avatars from a single portrait reference. This breakthrough exemplifies the potential of tri-plane representations in creating detailed and customizable 3D avatars. PET-NeuS by Wang et al. \cite{wang2023pet} extends the capabilities of the NeuS framework by incorporating a tri-plane representation borrowed from EG3D. By representing signed distance fields as a mixture of tri-planes and MLPs instead of using MLPs alone, the PET-NeuS method showcases an enhanced ability to process and visualize neural surfaces. Besides, TriHuman by Zhu et al. \cite{zhu2023trihuman} represents a novel human-tailored, deformable, and efficient tri-plane representation. Achieving real-time performance, state-of-the-art pose-controllable geometry synthesis, and photorealistic rendering quality.

\section{Important Datasets}
In this section, we compile a selection of notable datasets that have significantly contributed to advancements in the 3D field, encompassing, but not restricted to, the following examples:

\begin{itemize}

\item
\textbf{NYU Depth Dataset V2} \cite{silberman2012indoor} is a comprehensive collection of video sequences from various indoor scenes, captured using the RGB and Depth cameras of the Microsoft Kinect. It encompasses 1,449 densely labeled pairs of aligned RGB and depth images, 464 new scenes across three cities, and over 407,000 unlabeled frames. Each object within these images is meticulously labeled with a class and instance number. The dataset is divided into labeled and raw components, with the labeled subset containing synchronized and annotated pairs of RGB and depth frames, and the raw dataset providing the original RGB, depth, and accelerometer data. The labeled dataset also includes preprocessed depth maps with missing values filled in, and is available as a Matlab .mat file containing variables for accelerometer data, in-painted depth maps, RGB images, instance maps, object label masks, class names, and raw depth maps.

\item  
\textbf{ShapeNet} \cite{chang2015shapenet} is a comprehensive 3D CAD model repository, meticulously developed by a collaboration between Stanford University, Princeton University, and the Toyota Technological Institute at Chicago. It boasts over 300 million models, with 220,000 meticulously classified into 3,135 categories, following the WordNet hypernym-hyponym relationships for systematic organization. The ShapeNet Parts subset is particularly notable, encompassing 31,693 meshes across 16 common object classes, such as tables and chairs, with each model’s ground truth segmented into 2-5 parts, cumulatively spanning 50 part classes.

\item  
\textbf{ModelNet} \cite{wu20153d} is a comprehensive collection of 3D CAD models designed to support the research community in computer vision, graphics, robotics, and cognitive science. It consists of two subsets: ModelNet10 and ModelNet40. ModelNet10 is a curated set of CAD models from 10 popular object categories, with orientations manually aligned, while ModelNet40 expands this to 40 categories. The dataset was created by compiling a list of common object categories from the SUN database, followed by collecting CAD models for each category through online searches. Human workers on Amazon Mechanical Turk then manually verified the categorization of each model. ModelNet40 contains 12,311 CAD-generated meshes, with 9,843 designated for training and 2,468 for testing.

\item  
\textbf{ScanNet} \cite{dai2017scannet} is a richly-annotated dataset of 3D indoor spaces that provides over 1,500 scans captured with a Matterport camera in various buildings. Each scan in the dataset is a 3D reconstruction of an indoor scene containing a mesh in the PLY format, which includes vertex coordinates and colors. The dataset also features semantic and instance annotations, with over 20 object categories mapped to NYU40 labels. Annotations include axis-aligned bounding boxes for objects, and the dataset supports tasks like semantic segmentation, object detection, and scene completion. ScanNet’s data structure is organized into train and validation sets, with 1,201 training and 312 validation scans, each stored in its respective folder. The dataset preparation involves exporting the raw point cloud data and generating relevant annotations, including semantic labels, instance labels, and ground truth bounding boxes.

\item
\textbf{Matterport3D} \cite{chang2017matterport3d} is a comprehensive RGB-D dataset designed for indoor environments, featuring 10,800 panoramic views derived from 194,400 RGB-D images across 90 building-scale scenes. It provides detailed annotations including surface reconstructions, camera poses, and 2D and 3D semantic segmentations. Captured with the Matterport Pro 3D Camera, the dataset’s precise global alignment and diverse panoramic views facilitate a variety of computer vision tasks such as keypoint matching, view overlap prediction, normal prediction from color, semantic segmentation, and scene classification.

\item
\textbf{Pix3D} \cite{sun2018pix3d} is a dataset designed for single-image 3D shape modeling, offering a large-scale collection of image-shape pairs with precise pixel-level 2D-3D alignment. It includes a variety of object categories, each accompanied by high-quality images, segmentation masks, and corresponding 3D models. The dataset provides comprehensive metadata for each instance, such as the image path, object category, image size, 2D keypoints annotated by multiple individuals, and the object mask path. Additionally, it contains detailed 3D model information, including raw and voxelized models, 3D keypoints, and the source of the 3D model. Pix3D also specifies the camera parameters used in rendering, like rotation and translation matrices, focal length, camera position, and in-plane rotation, which are essential for tasks like reconstruction and viewpoint estimation.

\item  
\textbf{3DPW} \cite{von2018recovering} (3D Poses in the Wild), introduced by the Max Planck Society, is a pioneering collection designed for the evaluation of accurate 3D human poses in uncontrolled outdoor environments. Unlike previous datasets confined to limited recording volumes, 3DPW expands the scope by incorporating video footage captured from a mobile phone camera in motion. It encompasses 60 video sequences, complemented by 2D pose annotations and 3D poses derived through a method that combines video and Inertial Measurement Units (IMUs), ensuring high accuracy despite the scenes’ complexity. Additionally, the dataset provides camera poses for each frame, along with 3D body scans and reconfigurable 3D human models, featuring 18 distinct models in various clothing options.

\item  
\textbf{THUman} \cite{zheng2019deephuman}, specifically the THUman2.0 version, is a collection of 500 high-quality human scans, which were captured using a dense DSLR rig to ensure detailed and accurate representations. Each individual scan within the dataset is accompanied by a 3D model file in OBJ format and a corresponding texture map in JPEG format, providing the necessary data for realistic rendering. Additionally, the dataset has been updated to include SMPL-X fitting parameters and the corresponding meshes, which are available for download, allowing for advanced applications in human body shape and pose estimation.

\item
\textbf{ScanObjectNN} \cite{uy2019revisiting} is a benchmark dataset introduced in 2019, designed for point cloud classification challenges, featuring approximately 15,000 objects across 15 categories with 2,902 unique instances. The dataset stands out due to its real-world data, derived from indoor scans, presenting objects often obscured or partially visible, adding to the complexity of classification tasks. Each object in ScanObjectNN is represented by a point list, encompassing both global and local coordinates, alongside normals, color attributes, and semantic labels, making it a comprehensive resource for developing and testing 3D object recognition systems. Additionally, ScanObjectNN offers part annotations, a unique feature for real-world datasets, which aids in more granular analysis and model training. The dataset provides various variants like OBJ\_BG, PB\_T25, PB\_T25\_R, PB\_T50\_R, and PB\_T50\_RS, catering to different levels of background inclusion and object transformations, thus allowing researchers to rigorously evaluate the robustness of their classification models.

\item
\textbf{3D-Future} \cite{fu20213d} dataset is a comprehensive collection designed to bridge the gap between current 3D object modeling and industrial needs for 3D vision. It features  20,240 photo-realistic synthetic images  captured across  5,000 diverse scenes , rendered using advanced industrial 3D renders. The dataset includes  9,992 unique industrial 3D CAD shapes of furniture, each with high-resolution, informative textures developed by professional designers. These models showcase a variety of styles, including intricate European designs, and are enriched with different attribute labels to facilitate detailed research. The 3D-Future dataset, augmented to include an additional  6,571 3D furniture models , aims to support innovative research in texture recovery and transfer, high-quality 3D shape understanding, and generation, catering to the nuanced requirements of industrial production.

\item  
\textbf{ABO} \cite{collins2022abo} is a dataset designed to bridge the gap between real-world objects and their virtual 3D representations. It encompasses a wide array of product catalog images, detailed metadata, and high-resolution, artist-created 3D models that mirror the complex geometries and physically-based materials of actual household items. This dataset is particularly significant for its application in advancing the field of 3D object understanding, providing a robust platform for developing and testing algorithms across various challenges, including single-view 3D reconstruction, material estimation, and cross-domain multi-view object retrieval. The ABO dataset stands out for its realistic renderings and the inclusion of Blender renderings from 30 different viewpoints for each of the 7,953 objects it contains, complete with camera intrinsics and extrinsics for each rendering.

\item  
\textbf{Objaverse} \cite{deitke2023objaverse} is a comprehensive 3D dataset that stands out for its vast scale and diversity, encompassing over 800,000 high-quality 3D models that continue to grow in number. It is meticulously annotated with descriptive captions, tags, and animations, making it a versatile resource for various AI applications. The dataset’s extensive range of object categories and the visual variety within each category significantly surpasses existing 3D repositories. Objaverse’s potential is demonstrated through its application in training generative 3D models, enhancing segmentation in tail categories, aiding open-vocabulary object-navigation models in Embodied AI, and establishing robustness benchmarks for vision models.

\item
\textbf{Objaverse-XL} \cite{deitke2024objaverse} is a comprehensive dataset that encompasses over 10 million 3D objects, making it a significant resource for the field of 3D computer vision. It’s designed to facilitate the training of foundation models like Zero123-XL, which demonstrate remarkable 3D generalization abilities. The dataset is a diverse collection of deduplicated 3D objects sourced from various origins, including manually designed objects, photogrammetry scans of landmarks and everyday items, as well as professional scans of historic and antique artifacts. With its vast scale, Objaverse-XL is twelve times larger than its predecessor, Objaverse 1.0, and a hundred times larger than all other 3D datasets combined. This dataset is not only open for access but also comes with API scripts for downloading and processing, and it’s compatible with Blender for rendering.

\item  
\textbf{Cap3D} \cite{luo2024scalable} is a comprehensive 3D dataset designed for scalable 3D captioning, which was introduced at NeurIPS 2023. It includes over a million descriptive captions for 3D objects, specifically from the Objaverse and a subset of Objaverse-XL datasets. The dataset is enriched with 1,002,422 point clouds and rendered images, complete with camera, depth, and MatAlpha information corresponding to the objects with captions. The Cap3D dataset leverages pretrained models to provide detailed descriptions of 3D objects by consolidating multi-view information, thus facilitating an automatic approach for generating descriptive text for 3D objects and bypassing the need for manual annotation.

\end{itemize}

\section{Research Directions and Future Work}
In this section, we discuss research directions and the prospects of future advancements in the realm of 3D representation studies.
\begin{itemize}
\item Improved Efficiency: Developing more computationally efficient methods for real-time 3D representation and rendering, reducing the computational burden.
\item Hybrid Approaches: Exploring hybrid methods that combine the strengths of different representation techniques to handle diverse applications more effectively.
\item Scalability: Enhancing the scalability of 3D representations to manage larger and more complex scenes without compromising performance.
\item Interdisciplinary Applications: Applying 3D representation methods to new domains such as medical imaging, geospatial analysis, and cultural heritage preservation.
\item Data Augmentation and Generation: Leveraging generative models to create diverse and realistic 3D datasets, aiding in training and evaluation of 3D algorithms.
\item Integration with AR/VR: Advancing the integration of 3D representations in augmented reality (AR) and virtual reality (VR) environments for immersive experiences.
\item Deformable and Dynamic Representations: Researching and developing 3D representations that can dynamically adapt to changes, such as deformable objects or environments that evolve over time, enhancing realism and interactivity in simulations and virtual experiences.
\end{itemize}

\section{Conclusion}
This survey has provided a detailed exploration of the development, methodologies, and applications of various 3D representation methods. From traditional geometric models to cutting-edge neural representations, each approach offers unique advantages and faces distinct challenges. By introducing key datasets and identifying future research directions, this survey aims to facilitate ongoing and future efforts in the field.

{
    \small
    \bibliographystyle{ieeenat_fullname}
    \bibliography{main}

\begin{thebibliography}{59}
\providecommand{\natexlab}[1]{#1}
\providecommand{\url}[1]{\texttt{#1}}
\expandafter\ifx\csname urlstyle\endcsname\relax
  \providecommand{\doi}[1]{doi: #1}\else
  \providecommand{\doi}{doi: \begingroup \urlstyle{rm}\Url}\fi

\bibitem[Barron et~al.(2021)Barron, Mildenhall, Tancik, Hedman, Martin-Brualla, and Srinivasan]{barron2021mip}
Jonathan~T Barron, Ben Mildenhall, Matthew Tancik, Peter Hedman, Ricardo Martin-Brualla, and Pratul~P Srinivasan.
\newblock Mip-nerf: A multiscale representation for anti-aliasing neural radiance fields.
\newblock In \emph{Proceedings of the IEEE/CVF International Conference on Computer Vision}, pages 5855--5864, 2021.

\bibitem[Barron et~al.(2022)Barron, Mildenhall, Verbin, Srinivasan, and Hedman]{barron2022mip}
Jonathan~T Barron, Ben Mildenhall, Dor Verbin, Pratul~P Srinivasan, and Peter Hedman.
\newblock Mip-nerf 360: Unbounded anti-aliased neural radiance fields.
\newblock In \emph{Proceedings of the IEEE/CVF Conference on Computer Vision and Pattern Recognition}, pages 5470--5479, 2022.

\bibitem[Barron et~al.(2023)Barron, Mildenhall, Verbin, Srinivasan, and Hedman]{barron2023zip}
Jonathan~T Barron, Ben Mildenhall, Dor Verbin, Pratul~P Srinivasan, and Peter Hedman.
\newblock Zip-nerf: Anti-aliased grid-based neural radiance fields.
\newblock In \emph{Proceedings of the IEEE/CVF International Conference on Computer Vision}, pages 19697--19705, 2023.

\bibitem[Chan et~al.(2022)Chan, Lin, Chan, Nagano, Pan, De~Mello, Gallo, Guibas, Tremblay, Khamis, et~al.]{chan2022efficient}
Eric~R Chan, Connor~Z Lin, Matthew~A Chan, Koki Nagano, Boxiao Pan, Shalini De~Mello, Orazio Gallo, Leonidas~J Guibas, Jonathan Tremblay, Sameh Khamis, et~al.
\newblock Efficient geometry-aware 3d generative adversarial networks.
\newblock In \emph{Proceedings of the IEEE/CVF conference on computer vision and pattern recognition}, pages 16123--16133, 2022.

\bibitem[Chang et~al.(2017)Chang, Dai, Funkhouser, Halber, Niessner, Savva, Song, Zeng, and Zhang]{chang2017matterport3d}
Angel Chang, Angela Dai, Thomas Funkhouser, Maciej Halber, Matthias Niessner, Manolis Savva, Shuran Song, Andy Zeng, and Yinda Zhang.
\newblock Matterport3d: Learning from rgb-d data in indoor environments.
\newblock \emph{arXiv preprint arXiv:1709.06158}, 2017.

\bibitem[Chang et~al.(2015)Chang, Funkhouser, Guibas, Hanrahan, Huang, Li, Savarese, Savva, Song, Su, et~al.]{chang2015shapenet}
Angel~X Chang, Thomas Funkhouser, Leonidas Guibas, Pat Hanrahan, Qixing Huang, Zimo Li, Silvio Savarese, Manolis Savva, Shuran Song, Hao Su, et~al.
\newblock Shapenet: An information-rich 3d model repository.
\newblock \emph{arXiv preprint arXiv:1512.03012}, 2015.

\bibitem[Chen et~al.(2023)Chen, Chen, Jiao, and Jia]{chen2023fantasia3d}
Rui Chen, Yongwei Chen, Ningxin Jiao, and Kui Jia.
\newblock Fantasia3d: Disentangling geometry and appearance for high-quality text-to-3d content creation.
\newblock In \emph{Proceedings of the IEEE/CVF International Conference on Computer Vision}, pages 22246--22256, 2023.

\bibitem[Collins et~al.(2022)Collins, Goel, Deng, Luthra, Xu, Gundogdu, Zhang, Vicente, Dideriksen, Arora, et~al.]{collins2022abo}
Jasmine Collins, Shubham Goel, Kenan Deng, Achleshwar Luthra, Leon Xu, Erhan Gundogdu, Xi Zhang, Tomas F~Yago Vicente, Thomas Dideriksen, Himanshu Arora, et~al.
\newblock Abo: Dataset and benchmarks for real-world 3d object understanding.
\newblock In \emph{Proceedings of the IEEE/CVF conference on computer vision and pattern recognition}, pages 21126--21136, 2022.

\bibitem[Dai et~al.(2017)Dai, Chang, Savva, Halber, Funkhouser, and Nie{\ss}ner]{dai2017scannet}
Angela Dai, Angel~X Chang, Manolis Savva, Maciej Halber, Thomas Funkhouser, and Matthias Nie{\ss}ner.
\newblock Scannet: Richly-annotated 3d reconstructions of indoor scenes.
\newblock In \emph{Proceedings of the IEEE conference on computer vision and pattern recognition}, pages 5828--5839, 2017.

\bibitem[Deitke et~al.(2023)Deitke, Schwenk, Salvador, Weihs, Michel, VanderBilt, Schmidt, Ehsani, Kembhavi, and Farhadi]{deitke2023objaverse}
Matt Deitke, Dustin Schwenk, Jordi Salvador, Luca Weihs, Oscar Michel, Eli VanderBilt, Ludwig Schmidt, Kiana Ehsani, Aniruddha Kembhavi, and Ali Farhadi.
\newblock Objaverse: A universe of annotated 3d objects.
\newblock In \emph{Proceedings of the IEEE/CVF Conference on Computer Vision and Pattern Recognition}, pages 13142--13153, 2023.

\bibitem[Deitke et~al.(2024)Deitke, Liu, Wallingford, Ngo, Michel, Kusupati, Fan, Laforte, Voleti, Gadre, et~al.]{deitke2024objaverse}
Matt Deitke, Ruoshi Liu, Matthew Wallingford, Huong Ngo, Oscar Michel, Aditya Kusupati, Alan Fan, Christian Laforte, Vikram Voleti, Samir~Yitzhak Gadre, et~al.
\newblock Objaverse-xl: A universe of 10m+ 3d objects.
\newblock \emph{Advances in Neural Information Processing Systems}, 36, 2024.

\bibitem[Fridovich-Keil et~al.(2022)Fridovich-Keil, Yu, Tancik, Chen, Recht, and Kanazawa]{fridovich2022plenoxels}
Sara Fridovich-Keil, Alex Yu, Matthew Tancik, Qinhong Chen, Benjamin Recht, and Angjoo Kanazawa.
\newblock Plenoxels: Radiance fields without neural networks.
\newblock In \emph{Proceedings of the IEEE/CVF Conference on Computer Vision and Pattern Recognition}, pages 5501--5510, 2022.

\bibitem[Fu et~al.(2021)Fu, Jia, Gao, Gong, Zhao, Maybank, and Tao]{fu20213d}
Huan Fu, Rongfei Jia, Lin Gao, Mingming Gong, Binqiang Zhao, Steve Maybank, and Dacheng Tao.
\newblock 3d-future: 3d furniture shape with texture.
\newblock \emph{International Journal of Computer Vision}, 129:\penalty0 3313--3337, 2021.

\bibitem[Gezawa et~al.(2020)Gezawa, Zhang, Wang, and Yunqi]{gezawa2020review}
Abubakar~Sulaiman Gezawa, Yan Zhang, Qicong Wang, and Lei Yunqi.
\newblock A review on deep learning approaches for 3d data representations in retrieval and classifications.
\newblock \emph{IEEE access}, 8:\penalty0 57566--57593, 2020.

\bibitem[Gkioxari et~al.(2019)Gkioxari, Malik, and Johnson]{gkioxari2019mesh}
Georgia Gkioxari, Jitendra Malik, and Justin Johnson.
\newblock Mesh r-cnn.
\newblock In \emph{Proceedings of the IEEE/CVF international conference on computer vision}, pages 9785--9795, 2019.

\bibitem[Groueix et~al.(2018)Groueix, Fisher, Kim, Russell, and Aubry]{groueix2018papier}
Thibault Groueix, Matthew Fisher, Vladimir~G Kim, Bryan~C Russell, and Mathieu Aubry.
\newblock A papier-m{\^a}ch{\'e} approach to learning 3d surface generation.
\newblock In \emph{Proceedings of the IEEE conference on computer vision and pattern recognition}, pages 216--224, 2018.

\bibitem[Huang et~al.(2023)Huang, Zheng, Zhang, Zhou, and Lu]{huang2023tri}
Yuanhui Huang, Wenzhao Zheng, Yunpeng Zhang, Jie Zhou, and Jiwen Lu.
\newblock Tri-perspective view for vision-based 3d semantic occupancy prediction.
\newblock In \emph{Proceedings of the IEEE/CVF conference on computer vision and pattern recognition}, pages 9223--9232, 2023.

\bibitem[Jun and Nichol(2023)]{jun2023shap}
Heewoo Jun and Alex Nichol.
\newblock Shap-e: Generating conditional 3d implicit functions.
\newblock \emph{arXiv preprint arXiv:2305.02463}, 2023.

\bibitem[Kato et~al.(2018)Kato, Ushiku, and Harada]{kato2018neural}
Hiroharu Kato, Yoshitaka Ushiku, and Tatsuya Harada.
\newblock Neural 3d mesh renderer.
\newblock In \emph{Proceedings of the IEEE conference on computer vision and pattern recognition}, pages 3907--3916, 2018.

\bibitem[Kerbl et~al.(2023)Kerbl, Kopanas, Leimk{\"u}hler, and Drettakis]{kerbl20233d}
Bernhard Kerbl, Georgios Kopanas, Thomas Leimk{\"u}hler, and George Drettakis.
\newblock 3d gaussian splatting for real-time radiance field rendering.
\newblock \emph{ACM Transactions on Graphics}, 42\penalty0 (4):\penalty0 1--14, 2023.

\bibitem[Lin et~al.(2023)Lin, Gao, Tang, Takikawa, Zeng, Huang, Kreis, Fidler, Liu, and Lin]{lin2023magic3d}
Chen-Hsuan Lin, Jun Gao, Luming Tang, Towaki Takikawa, Xiaohui Zeng, Xun Huang, Karsten Kreis, Sanja Fidler, Ming-Yu Liu, and Tsung-Yi Lin.
\newblock Magic3d: High-resolution text-to-3d content creation.
\newblock In \emph{Proceedings of the IEEE/CVF Conference on Computer Vision and Pattern Recognition}, pages 300--309, 2023.

\bibitem[Luiten et~al.(2023)Luiten, Kopanas, Leibe, and Ramanan]{luiten2023dynamic}
Jonathon Luiten, Georgios Kopanas, Bastian Leibe, and Deva Ramanan.
\newblock Dynamic 3d gaussians: Tracking by persistent dynamic view synthesis.
\newblock \emph{arXiv preprint arXiv:2308.09713}, 2023.

\bibitem[Luo et~al.(2024)Luo, Rockwell, Lee, and Johnson]{luo2024scalable}
Tiange Luo, Chris Rockwell, Honglak Lee, and Justin Johnson.
\newblock Scalable 3d captioning with pretrained models.
\newblock \emph{Advances in Neural Information Processing Systems}, 36, 2024.

\bibitem[Ma et~al.(2023)Ma, Zhu, Qi, Lei, and Zhang]{ma2023otavatar}
Zhiyuan Ma, Xiangyu Zhu, Guo-Jun Qi, Zhen Lei, and Lei Zhang.
\newblock Otavatar: One-shot talking face avatar with controllable tri-plane rendering.
\newblock In \emph{Proceedings of the IEEE/CVF Conference on Computer Vision and Pattern Recognition}, pages 16901--16910, 2023.

\bibitem[Martin-Brualla et~al.(2021)Martin-Brualla, Radwan, Sajjadi, Barron, Dosovitskiy, and Duckworth]{martin2021nerf}
Ricardo Martin-Brualla, Noha Radwan, Mehdi~SM Sajjadi, Jonathan~T Barron, Alexey Dosovitskiy, and Daniel Duckworth.
\newblock Nerf in the wild: Neural radiance fields for unconstrained photo collections.
\newblock In \emph{Proceedings of the IEEE/CVF Conference on Computer Vision and Pattern Recognition}, pages 7210--7219, 2021.

\bibitem[Maturana and Scherer(2015)]{maturana2015voxnet}
Daniel Maturana and Sebastian Scherer.
\newblock Voxnet: A 3d convolutional neural network for real-time object recognition.
\newblock In \emph{2015 IEEE/RSJ international conference on intelligent robots and systems (IROS)}, pages 922--928. IEEE, 2015.

\bibitem[Mildenhall et~al.(2021)Mildenhall, Srinivasan, Tancik, Barron, Ramamoorthi, and Ng]{mildenhall2021nerf}
Ben Mildenhall, Pratul~P Srinivasan, Matthew Tancik, Jonathan~T Barron, Ravi Ramamoorthi, and Ren Ng.
\newblock Nerf: Representing scenes as neural radiance fields for view synthesis.
\newblock \emph{Communications of the ACM}, 65\penalty0 (1):\penalty0 99--106, 2021.

\bibitem[Mittal et~al.(2022)Mittal, Cheng, Singh, and Tulsiani]{mittal2022autosdf}
Paritosh Mittal, Yen-Chi Cheng, Maneesh Singh, and Shubham Tulsiani.
\newblock Autosdf: Shape priors for 3d completion, reconstruction and generation.
\newblock In \emph{Proceedings of the IEEE/CVF Conference on Computer Vision and Pattern Recognition}, pages 306--315, 2022.

\bibitem[Mu et~al.(2021)Mu, Qiu, Kortylewski, Yuille, Vasconcelos, and Wang]{mu2021sdf}
Jiteng Mu, Weichao Qiu, Adam Kortylewski, Alan Yuille, Nuno Vasconcelos, and Xiaolong Wang.
\newblock A-sdf: Learning disentangled signed distance functions for articulated shape representation.
\newblock In \emph{Proceedings of the IEEE/CVF International Conference on Computer Vision}, pages 13001--13011, 2021.

\bibitem[M{\"u}ller et~al.(2022)M{\"u}ller, Evans, Schied, and Keller]{muller2022instant}
Thomas M{\"u}ller, Alex Evans, Christoph Schied, and Alexander Keller.
\newblock Instant neural graphics primitives with a multiresolution hash encoding.
\newblock \emph{ACM transactions on graphics (TOG)}, 41\penalty0 (4):\penalty0 1--15, 2022.

\bibitem[Nichol et~al.(2022)Nichol, Jun, Dhariwal, Mishkin, and Chen]{nichol2022point}
Alex Nichol, Heewoo Jun, Prafulla Dhariwal, Pamela Mishkin, and Mark Chen.
\newblock Point-e: A system for generating 3d point clouds from complex prompts.
\newblock \emph{arXiv preprint arXiv:2212.08751}, 2022.

\bibitem[Park et~al.(2019)Park, Florence, Straub, Newcombe, and Lovegrove]{park2019deepsdf}
Jeong~Joon Park, Peter Florence, Julian Straub, Richard Newcombe, and Steven Lovegrove.
\newblock Deepsdf: Learning continuous signed distance functions for shape representation.
\newblock In \emph{Proceedings of the IEEE/CVF conference on computer vision and pattern recognition}, pages 165--174, 2019.

\bibitem[Qi et~al.(2017{\natexlab{a}})Qi, Su, Mo, and Guibas]{qi2017pointnet}
Charles~R Qi, Hao Su, Kaichun Mo, and Leonidas~J Guibas.
\newblock Pointnet: Deep learning on point sets for 3d classification and segmentation.
\newblock In \emph{Proceedings of the IEEE conference on computer vision and pattern recognition}, pages 652--660, 2017{\natexlab{a}}.

\bibitem[Qi et~al.(2017{\natexlab{b}})Qi, Yi, Su, and Guibas]{qi2017pointnet++}
Charles~Ruizhongtai Qi, Li Yi, Hao Su, and Leonidas~J Guibas.
\newblock Pointnet++: Deep hierarchical feature learning on point sets in a metric space.
\newblock \emph{Advances in neural information processing systems}, 30, 2017{\natexlab{b}}.

\bibitem[Schwarz et~al.(2020)Schwarz, Liao, Niemeyer, and Geiger]{schwarz2020graf}
Katja Schwarz, Yiyi Liao, Michael Niemeyer, and Andreas Geiger.
\newblock Graf: Generative radiance fields for 3d-aware image synthesis.
\newblock \emph{Advances in Neural Information Processing Systems}, 33:\penalty0 20154--20166, 2020.

\bibitem[Schwarz et~al.(2022)Schwarz, Sauer, Niemeyer, Liao, and Geiger]{schwarz2022voxgraf}
Katja Schwarz, Axel Sauer, Michael Niemeyer, Yiyi Liao, and Andreas Geiger.
\newblock Voxgraf: Fast 3d-aware image synthesis with sparse voxel grids.
\newblock \emph{Advances in Neural Information Processing Systems}, 35:\penalty0 33999--34011, 2022.

\bibitem[Sella et~al.(2023)Sella, Fiebelman, Hedman, and Averbuch-Elor]{sella2023vox}
Etai Sella, Gal Fiebelman, Peter Hedman, and Hadar Averbuch-Elor.
\newblock Vox-e: Text-guided voxel editing of 3d objects.
\newblock In \emph{Proceedings of the IEEE/CVF International Conference on Computer Vision}, pages 430--440, 2023.

\bibitem[Shen et~al.(2021)Shen, Gao, Yin, Liu, and Fidler]{shen2021deep}
Tianchang Shen, Jun Gao, Kangxue Yin, Ming-Yu Liu, and Sanja Fidler.
\newblock Deep marching tetrahedra: a hybrid representation for high-resolution 3d shape synthesis.
\newblock \emph{Advances in Neural Information Processing Systems}, 34:\penalty0 6087--6101, 2021.

\bibitem[Silberman et~al.(2012)Silberman, Hoiem, Kohli, and Fergus]{silberman2012indoor}
Nathan Silberman, Derek Hoiem, Pushmeet Kohli, and Rob Fergus.
\newblock Indoor segmentation and support inference from rgbd images.
\newblock In \emph{Computer Vision--ECCV 2012: 12th European Conference on Computer Vision, Florence, Italy, October 7-13, 2012, Proceedings, Part V 12}, pages 746--760. Springer, 2012.

\bibitem[Sun et~al.(2018)Sun, Wu, Zhang, Zhang, Zhang, Xue, Tenenbaum, and Freeman]{sun2018pix3d}
Xingyuan Sun, Jiajun Wu, Xiuming Zhang, Zhoutong Zhang, Chengkai Zhang, Tianfan Xue, Joshua~B Tenenbaum, and William~T Freeman.
\newblock Pix3d: Dataset and methods for single-image 3d shape modeling.
\newblock In \emph{Proceedings of the IEEE conference on computer vision and pattern recognition}, pages 2974--2983, 2018.

\bibitem[Tang et~al.(2023)Tang, Ren, Zhou, Liu, and Zeng]{tang2023dreamgaussian}
Jiaxiang Tang, Jiawei Ren, Hang Zhou, Ziwei Liu, and Gang Zeng.
\newblock Dreamgaussian: Generative gaussian splatting for efficient 3d content creation.
\newblock \emph{arXiv preprint arXiv:2309.16653}, 2023.

\bibitem[Tatarchenko et~al.(2017)Tatarchenko, Dosovitskiy, and Brox]{tatarchenko2017octree}
Maxim Tatarchenko, Alexey Dosovitskiy, and Thomas Brox.
\newblock Octree generating networks: Efficient convolutional architectures for high-resolution 3d outputs.
\newblock In \emph{Proceedings of the IEEE international conference on computer vision}, pages 2088--2096, 2017.

\bibitem[Uy et~al.(2019)Uy, Pham, Hua, Nguyen, and Yeung]{uy2019revisiting}
Mikaela~Angelina Uy, Quang-Hieu Pham, Binh-Son Hua, Thanh Nguyen, and Sai-Kit Yeung.
\newblock Revisiting point cloud classification: A new benchmark dataset and classification model on real-world data.
\newblock In \emph{Proceedings of the IEEE/CVF international conference on computer vision}, pages 1588--1597, 2019.

\bibitem[Von~Marcard et~al.(2018)Von~Marcard, Henschel, Black, Rosenhahn, and Pons-Moll]{von2018recovering}
Timo Von~Marcard, Roberto Henschel, Michael~J Black, Bodo Rosenhahn, and Gerard Pons-Moll.
\newblock Recovering accurate 3d human pose in the wild using imus and a moving camera.
\newblock In \emph{Proceedings of the European conference on computer vision (ECCV)}, pages 601--617, 2018.

\bibitem[Wang et~al.(2018)Wang, Zhang, Li, Fu, Liu, and Jiang]{wang2018pixel2mesh}
Nanyang Wang, Yinda Zhang, Zhuwen Li, Yanwei Fu, Wei Liu, and Yu-Gang Jiang.
\newblock Pixel2mesh: Generating 3d mesh models from single rgb images.
\newblock In \emph{Proceedings of the European conference on computer vision (ECCV)}, pages 52--67, 2018.

\bibitem[Wang et~al.(2019)Wang, Sun, Liu, Sarma, Bronstein, and Solomon]{wang2019dynamic}
Yue Wang, Yongbin Sun, Ziwei Liu, Sanjay~E Sarma, Michael~M Bronstein, and Justin~M Solomon.
\newblock Dynamic graph cnn for learning on point clouds.
\newblock \emph{ACM Transactions on Graphics (tog)}, 38\penalty0 (5):\penalty0 1--12, 2019.

\bibitem[Wang et~al.(2023)Wang, Skorokhodov, and Wonka]{wang2023pet}
Yiqun Wang, Ivan Skorokhodov, and Peter Wonka.
\newblock Pet-neus: Positional encoding tri-planes for neural surfaces.
\newblock In \emph{Proceedings of the IEEE/CVF Conference on Computer Vision and Pattern Recognition}, pages 12598--12607, 2023.

\bibitem[Wang et~al.(2021)Wang, Wu, Xie, Chen, and Prisacariu]{wang2021nerf}
Zirui Wang, Shangzhe Wu, Weidi Xie, Min Chen, and Victor~Adrian Prisacariu.
\newblock Nerf--: Neural radiance fields without known camera parameters.
\newblock \emph{arXiv preprint arXiv:2102.07064}, 2021.

\bibitem[Wu et~al.(2023)Wu, Yi, Fang, Xie, Zhang, Wei, Liu, Tian, and Wang]{wu20234d}
Guanjun Wu, Taoran Yi, Jiemin Fang, Lingxi Xie, Xiaopeng Zhang, Wei Wei, Wenyu Liu, Qi Tian, and Xinggang Wang.
\newblock 4d gaussian splatting for real-time dynamic scene rendering.
\newblock \emph{arXiv preprint arXiv:2310.08528}, 2023.

\bibitem[Wu et~al.(2015)Wu, Song, Khosla, Yu, Zhang, Tang, and Xiao]{wu20153d}
Zhirong Wu, Shuran Song, Aditya Khosla, Fisher Yu, Linguang Zhang, Xiaoou Tang, and Jianxiong Xiao.
\newblock 3d shapenets: A deep representation for volumetric shapes.
\newblock In \emph{Proceedings of the IEEE conference on computer vision and pattern recognition}, pages 1912--1920, 2015.

\bibitem[Yang et~al.(2023)Yang, Gao, Zhou, Jiao, Zhang, and Jin]{yang2023deformable}
Ziyi Yang, Xinyu Gao, Wen Zhou, Shaohui Jiao, Yuqing Zhang, and Xiaogang Jin.
\newblock Deformable 3d gaussians for high-fidelity monocular dynamic scene reconstruction.
\newblock \emph{arXiv preprint arXiv:2309.13101}, 2023.

\bibitem[Yi et~al.(2023)Yi, Fang, Wu, Xie, Zhang, Liu, Tian, and Wang]{yi2023gaussiandreamer}
Taoran Yi, Jiemin Fang, Guanjun Wu, Lingxi Xie, Xiaopeng Zhang, Wenyu Liu, Qi Tian, and Xinggang Wang.
\newblock Gaussiandreamer: Fast generation from text to 3d gaussian splatting with point cloud priors.
\newblock \emph{arXiv preprint arXiv:2310.08529}, 2023.

\bibitem[Yu et~al.(2023)Yu, Chen, Huang, Sattler, and Geiger]{yu2023mip}
Zehao Yu, Anpei Chen, Binbin Huang, Torsten Sattler, and Andreas Geiger.
\newblock Mip-splatting: Alias-free 3d gaussian splatting.
\newblock \emph{arXiv preprint arXiv:2311.16493}, 2023.

\bibitem[Zhang et~al.(2020)Zhang, Riegler, Snavely, and Koltun]{zhang2020nerf++}
Kai Zhang, Gernot Riegler, Noah Snavely, and Vladlen Koltun.
\newblock Nerf++: Analyzing and improving neural radiance fields.
\newblock \emph{arXiv preprint arXiv:2010.07492}, 2020.

\bibitem[Zhao et~al.(2021)Zhao, Jiang, Jia, Torr, and Koltun]{zhao2021point}
Hengshuang Zhao, Li Jiang, Jiaya Jia, Philip~HS Torr, and Vladlen Koltun.
\newblock Point transformer.
\newblock In \emph{Proceedings of the IEEE/CVF international conference on computer vision}, pages 16259--16268, 2021.

\bibitem[Zheng et~al.(2022)Zheng, Liu, Wang, and Tong]{zheng2022sdf}
Xinyang Zheng, Yang Liu, Pengshuai Wang, and Xin Tong.
\newblock Sdf-stylegan: Implicit sdf-based stylegan for 3d shape generation.
\newblock In \emph{Computer Graphics Forum}, pages 52--63. Wiley Online Library, 2022.

\bibitem[Zheng et~al.(2023)Zheng, Pan, Wang, Tong, Liu, and Shum]{zheng2023locally}
Xin-Yang Zheng, Hao Pan, Peng-Shuai Wang, Xin Tong, Yang Liu, and Heung-Yeung Shum.
\newblock Locally attentional sdf diffusion for controllable 3d shape generation.
\newblock \emph{ACM Transactions on Graphics (TOG)}, 42\penalty0 (4):\penalty0 1--13, 2023.

\bibitem[Zheng et~al.(2019)Zheng, Yu, Wei, Dai, and Liu]{zheng2019deephuman}
Zerong Zheng, Tao Yu, Yixuan Wei, Qionghai Dai, and Yebin Liu.
\newblock Deephuman: 3d human reconstruction from a single image.
\newblock In \emph{Proceedings of the IEEE/CVF International Conference on Computer Vision}, pages 7739--7749, 2019.

\bibitem[Zhu et~al.(2023)Zhu, Zhan, Theobalt, and Habermann]{zhu2023trihuman}
Heming Zhu, Fangneng Zhan, Christian Theobalt, and Marc Habermann.
\newblock Trihuman: A real-time and controllable tri-plane representation for detailed human geometry and appearance synthesis.
\newblock \emph{arXiv preprint arXiv:2312.05161}, 2023.

\end{thebibliography}
}

\end{document}